# Analog Resonant Recurrent Neural Networks via Metacircuits

Zixin Zhou[1,3], Tianxi Jiang[1,3,*], Menglong Yang[1], Mengli Sui[1], Yizhuo Zhang[1], Zhihua Feng[1], Qingbo He[2,*], and Shiwu Zhang[1,*]

[1]Institute of Humanoid Robots, CAS Key Laboratory of Mechanical Behavior and Design of Materials, Department of Precision Machinery and Precision Instrumentation, University of Science and Technology of China, Hefei 230026, People's Republic of China

[2]State Key Laboratory of Mechanical System and Vibration, Shanghai Jiao Tong University, Shanghai 200240, People's Republic of China

[3]These authors contributed equally to this work.

*Corresponding author: jiangtx@ustc.edu.cn, qbhe@sjtu.edu.cn, and swzhang@ustc.edu.cn

## Abstract

Physical neural networks offer a promising route to fast and energy-efficient edge intelligence by embedding computation in the intrinsic dynamics of materials and devices. However, processing temporal signals requires recurrent networks with memory, whose trained dynamics are hard to reproduce faithfully in physical systems. Here we present an analog resonant recurrent neural network ($R^2$NN) implemented as a metacircuit of coupled local resonators. A reformulated mechanical–electrical analogy based on frequency-dependent negative resistances maps a trained network one-to-one onto the metacircuit, with every trainable parameter realized as a resistance. Training tunes local resonances and global resistive couplings so that past responses persist as recurrent memory and interfere with incoming inputs, and the resulting impedance landscape routes different frequencies to distinct outputs. Exploiting this recurrent memory, the metacircuit solves a temporal-parity task whose output depends on the entire input history. Building on the learned frequency selectivity, stand-alone $R^2$NN systems classify raw tactile, acoustic and vibration signals in real time and directly drive downstream hardware. This work establishes engineered locally resonant dynamics as a physical substrate for recurrent neural computation and opens a route to compact, low-latency neural hardware for machine intelligence at the edge.

## Introduction

As artificial intelligence (AI) extends into the physical world, information processing is moving from the cloud to the edge, enabling real-time intelligent services with improved privacy[1,2]. In applications such as wearable health monitoring[3,4], human–machine interaction[5,6] and industrial diagnostics[7], edge devices must process continuous analog data streams from physical sensors. Yet von Neumann architectures are constrained by the "memory wall" and "power wall": shuttling data between separate processing and memory units limits both speed and energy efficiency. Processing analog signals in the digital domain further requires analog-to-digital conversion, domain transformation and feature extraction; each step adds latency, power consumption and hardware overhead.

Physical neural networks (PNNs), which perform neural-network operations with the intrinsic responses of physical systems, have emerged as a promising paradigm for addressing these limitations[8-14]. Analog platforms based on memristors[1,15-17], photonic and phononic chips[18,19] and metamaterials[20-26] have demonstrated advantages in energy efficiency and throughput over conventional architectures in tasks such as image and speech recognition. Among these, wave-based platforms are particularly attractive for their inherent parallelism, broad bandwidth and low latency[27-30]. To date, however, most of these systems implement feedforward architectures suited to spatial or quasi-static inputs, such as fully connected, diffractive and convolutional networks[20-22,26,31]. Lacking internal state memory, they typically require continuous signals such as speech or vibration to be pre-encoded or transformed into another domain. Physical reservoir computing exploits the nonlinear dynamics and short-term memory of physical systems[32], but the reservoir itself is untrained, so its dynamics are not tailored to the task, and the task-specific computation is delegated to a trained digital readout, often with pre-encoded inputs.

Recurrent neural networks (RNNs), whose outputs depend on both current inputs and past states, are naturally suited to such sequential data. Wave dynamics can be mapped onto RNN computation[33-35], allowing temporal information to be stored in the dynamic response of a physical system. For low-frequency signals from human motion, speech or mechanical vibration, however, wave-scattering designs face a fundamental limitation: the wavelengths often exceed the dimensions of practical devices. Resonance, a phenomenon ubiquitous in physical and biological systems[24,36], provides a natural route around this constraint. Coupled resonators with different lifetimes can implement analog

RNNs with both short- and long-term memory[34], and metamaterials with tailored local resonances can classify low-frequency temporal sequences[35]. Yet such analog RNNs remain largely theoretical: multimodal structural resonances and sensitivity to fabrication variations create a simulation-to-reality gap that has hindered their experimental realization. Existing designs are also confined to a single physical domain, which limits their ability to process signals from diverse physical fields. Their integration with electromechanical systems, essential for practical edge computing, remains unexplored.

In this Article, we present an analog resonant recurrent neural network ($R^2$NN) implemented as a metacircuit of coupled electrical local resonators. A reformulated mechanical–electrical analogy (MEA), realized with generalized impedance converters (GICs), maps the $R^2$NN dynamics onto the metacircuit, with every trainable parameter implemented as a precisely settable resistance. The local resonance of each unit cell gives rise to an effective frequency-dependent negative resistance (FDNR) whose impedance changes sharply near resonance. Training tunes these local resonances together with the resistive couplings between cells, so that responses to earlier inputs persist and interfere with later ones, providing recurrent memory, and the resulting impedance landscape channels currents of different frequencies into distinct output cells. Because both mechanisms arise from the metacircuit dynamics, raw sensor signals enter the network without pre-encoding or digital feature extraction, and all trained computation takes place in the analog domain. The class is then read out by comparing output energies, without trained parameters. Experiments on a temporal-parity task demonstrate that the recurrent dynamics support history-dependent computation, and simulations show that nonlinear resonators extend this capability to inputs without explicit temporal cues. Exploiting the frequency selectivity shaped by training, stand-alone $R^2$NN systems further classify raw signals across physical domains in tactile perception, voice-command classification and drone condition monitoring.

## Results

### Concept of metacircuit-based $R^2$NN.

To physically implement the $R^2$NN, we construct a metacircuit composed of coupled local resonators based on FDNRs. The conventional admittance-type MEA maps force $f$ to current $i$, so that a mechanical local resonator (masses $M$ and $m$, stiffnesses $k_n$ and $k_c$, effective mass $M_{eff}$) maps onto an inductor-capacitor (LC) local resonator **(Fig. 1a)**. Such LC circuits, however, cannot be realized with the precision that a trained $R^2$NN requires: commercial components are restricted to discrete E-series values, and low-frequency inductors are bulky and imprecise. We therefore eliminate inductors by introducing GIC-based FDNRs (**Supplementary Note 1**), so that all trainable parameters can be realized with resistors, whose values are set precisely through series combinations. Because GIC-based elements must be grounded, we divide all impedances by $j\omega$ (the Bruton transformation[37]), which converts grounded capacitors into FDNR elements *D*, floating inductors into resistors *R*, and resistors into capacitors *C* (**Supplementary Note 2**). For a parallel *D*–*C*–*R* configuration driven by a current *i*,

$$D\frac{d^2u}{dt^2} + C\frac{du}{dt} + R^{-1}u = i, \quad (1)$$

which establishes the correspondences $i \sim f$, $u \sim x$, $D \sim m$, $C \sim c$ and $R^{-1} \sim k$, where *u* is the node voltage, *x* the displacement and *c* the damping coefficient (**Fig. 1b**). Because stiffness maps onto conductance, all trainable parameters of the $R^2$NN are realized by resistors, whereas the FDNR elements are not trained. An FDNR-based local resonator is accordingly built from the FDNR elements $D_M$, $D_m$, an internal resistor $R_n$ and coupling resistors $R_c$, and its macroscopic response is captured by an effective FDNR element (**Fig. 1c**)

$$D_{\mathrm{eff}} = D_M + \frac{D_m}{1-\left(\frac{\omega}{\omega_0}\right)^2}, \quad (2)$$

where $\omega_0 = 1/\sqrt{D_m R_n}$ is the local resonance frequency and $\omega_1 = \omega_0\sqrt{(D_M + D_m)/D_M}$ the out-of-phase resonance frequency (see **Supplementary Note 3** for the unit cell with damping). The effective impedance of the unit cell is

$$Z_{\mathrm{eff}} = -\frac{1}{\omega^2 D_{\mathrm{eff}}} = -\frac{\omega^2-\omega_0^2}{\omega^2(\omega^2-\omega_1^2)D_M}. \quad (3)$$

The unit cell is thus strongly frequency dependent (**Fig. 1d**). For $\omega < \omega_0$ it behaves as a positive FDNR, the electrical counterpart of a positive effective mass. As $\omega$ approaches $\omega_0$, $D_{\mathrm{eff}}$ diverges and $Z_{\mathrm{eff}}$ vanishes, so the cell becomes a current sink that diverts current from neighboring branches, while the internal oscillator response is amplified by $\beta = U_{\mathrm{out}}/U_{\mathrm{in}} = 1/[1-(\omega/\omega_0)^2]$. For $\omega_0 < \omega < \omega_1$, $D_{\mathrm{eff}}$ is negative, corresponding to a negative effective mass, and near $\omega_1$ the cell enters a high-impedance state. A fabricated unit cell shows a valley-shaped impedance magnitude with a minimum near $\omega_0$ and a sign reversal of its real part, with additional imaginary components introduced by parasitic capacitance (**Methods; Extended Data Fig. 1**).

When unit cells are coupled into a network, responses to earlier inputs persist in the coupled resonators and interfere with later ones, giving the network an intrinsic recurrent memory, while the frequency selectivity of each cell governs how signal energy is distributed among branches. Both mechanisms are set by the trained resistances $R_n$ and $R_c$. On this basis, we formulate the metacircuit as the electrical analog of a locally resonant metamaterial and establish its equivalence to RNN dynamics (**Fig. 1e**). Discretizing the undamped form of **equation (1)** with time step $\Delta t$ gives

$$\begin{bmatrix} \mathbf{u}_{t+1} \\ \mathbf{u}_t \end{bmatrix} = \begin{bmatrix} \mathbf{2} - \Delta t^2 \mathbf{D}^{-1}\mathbf{Y} & -\mathbf{1} \\ \mathbf{1} & \mathbf{0} \end{bmatrix} \begin{bmatrix} \mathbf{u}_t \\ \mathbf{u}_{t-1} \end{bmatrix} + \begin{bmatrix} \Delta t^2 \mathbf{D}^{-1} \\ \mathbf{0} \end{bmatrix} \mathbf{i}_t, \tag{4}$$

where $\mathbf{D}$ contains $D_M$ and $D_m$, and the admittance matrix $\mathbf{Y}$ is assembled from $R_n^{-1}$ and $R_c^{-1}$ according to the metacircuit topology (**Methods**). This is the hidden-state update of an RNN,

$$\mathbf{h}_t = \mathbf{W}^{\mathrm{h}}(\mathbf{Y})\mathbf{h}_{t-1} + \mathbf{W}^{i}\mathbf{i}_t, \tag{5}$$

with $\mathbf{h}_t = [\mathbf{u}_{t+1}, \mathbf{u}_t]^{\mathrm{T}}$. The output of channel $\ell$ is $\mathrm{y}_{\ell,t} = |(\mathbf{W}^{\mathrm{y}}\mathbf{h}_t)_\ell|^2$, where $\mathbf{W}^{y}$ extracts the internal-oscillator responses ($U_{\mathrm{out}}$) of designated output unit cells. With linear unit cells, the state update is linear and nonlinearity enters only through the intensity readout, whereas nonlinear local resonators make the state update itself nonlinear (**Supplementary Note 3**). Classification compares the time-integrated outputs, which correspond to the oscillation energy accumulated at the output cells:

$$E_{\mathrm{prb},\ell} = \textstyle\sum_t y_{\ell,t}, \quad \ell^* = \arg\max_\ell E_{\mathrm{prb},\ell}, \tag{6}$$

and the normalized energies serve as class probabilities in training (**Methods**).

The $R^2$NN is trained offline with a uniform damping added to equation (4), matching the decaying responses of the circuit, and the optimized parameters are mapped onto $R_n$ and $R_c$ (**Methods**). The fabricated metacircuit performs inference in continuous time: the input signal is injected as a current and processed by the metacircuit, which forms the recurrent core, and a microcontroller (MCU) samples the output oscillators, computes $E_{\mathrm{prb},\ell}$ and outputs a logic-level label for $\ell^*$ (**Methods; Extended Data Fig. 2; Supplementary Note 4**). All computation involving trained parameters therefore resides in the physical network, whereas the digital readout only compares output energies and contains no trainable parameters. This distinguishes the $R^2$NN from physical reservoir computing, where a fixed reservoir relies on a trained digital readout. The metacircuit-based $R^2$NN thereby processes dynamic physical signals directly through analog recurrent dynamics, supporting embedded intelligent sensing applications (**Fig. 1f**).

## Mechanism of the metacircuit-based $R^2$NN

### 1. Recurrent dynamics of $R^2$NN

To probe the recurrent dynamics of the $R^2$NN, we use a three-bit temporal-parity task in which three consecutive sinusoidal segments encode three bits and the class is whether the number of 1s is odd or even **(Extended Data Fig. 3a, b)**. Because the label depends jointly on all three bits, no individual bit or pair of bits determines the class. Bits 0 and 1 are encoded at 40 and 60 Hz (phase-neutral encoding) or at 40 and 62 Hz (phase-accumulating encoding), and consecutive segments are joined with continuous phase, so that the waveform has no phase jump at segment boundaries. We train linear and nonlinear $R^2$NNs as mechanical models on a 7 × 7 lattice and read the class from the two probes (**Extended Data Fig. 3c; Methods**). In the nonlinear $R^2$NN, the local resonators are Duffing oscillators with $k_{\mathrm{n}}^{\mathrm{non}} = k_{\mathrm{n}}^{\mathrm{linear}} + k_{\mathrm{cubic}} q^2$ (electrical counterpart $R_{\mathrm{n}}^{\mathrm{non}} = (k_{\mathrm{n}}^{\mathrm{non}})^{-1}$ ; **Supplementary Note 3**), where $q$ is the displacement of the internal mass relative to its primary mass. Because nonlinear elements are not yet available in our hardware, we study the nonlinear $R^2$NN numerically and verify the linear $R^2$NN experimentally.

A linear $R^2$NN reaches only 87.5% test accuracy with phase-neutral inputs but 100% with phase-accumulating inputs **(Extended Data Fig. 3e, f)**. The 40- and 60-Hz segments complete integer numbers of cycles, whereas each 62-Hz segment leaves a residual phase of 0.8π rad, so the phase of later segments records how many 1s have already occurred **(Extended Data Fig. 3g)**. Without retraining, the same model reaches only 62.5% when the source phase is reset at each segment boundary and 76.75% on phase-neutral inputs (**Extended Data Fig. 3h**). Without the phase channel, linear dynamics followed by a quadratic readout can express only single-bit and pairwise terms, which bounds the accuracy over the eight equally weighted patterns at 7/8 (87.5%; **Supplementary Note 5**). The linear $R^2$NN therefore solves parity by converting the history written into the input phase into probe-specific interference.

We map the 5 × 5 linear $R^2$NN trained with phase-accumulating encoding (simulated results in **Supplementary Note 6**) onto a metacircuit and test it experimentally **(Fig. 2; Methods)**. For all eight patterns, the dominant probe at the end of the sequence matches the parity label and holds 0.60–0.84 of the total normalized energy (**Fig. 2a**). Across 50 noisy waveforms per pattern, the energy difference forms pattern-specific clusters on the correct side of the decision boundary (**Fig. 2b**). The $R^2$NN classifies all 400 recordings correctly (100%), whether its probe energies are compared offline from the recorded voltages or by the MCU in real time **(Fig. 2c)**. Its robustness, including the conditions under which accuracy degrades, is characterized in **Supplementary Note 7**. Band-pass filters and isolated unit cells tuned to the two carriers fall well short of this accuracy with the same readout (**Supplementary Note 8**), so the 100% accuracy stems from the trained coupled dynamics rather than from carrier selection. The readout is a simple running sum of the squared probe voltages, computed while the input is applied, so the decision is available within 0.1 ms after the input ends. The power, latency, energy and component count of the system are reported in **Supplementary Note 9**.

With phase-neutral encoding, a nonlinear 7 × 7 $R^2$NN reaches 100% test accuracy (**Extended Data Fig. 3d**). State retention alone does not explain this result, because the intact model and its $k_{\mathrm{cubic}}$ = 0 counterpart retain comparable responses after the input stops (**Extended Data Fig. 3i, left**). Instead, in the intact model the response to the full input deviates by 55.6% from the sum of the responses to the history and to the current segment applied separately, whereas this non-additivity vanishes when $k_{\mathrm{cubic}}$ = 0 (**Extended Data Fig. 3i, right**). The stored state thus reshapes the response to new input rather than merely adding to it: switching the third bit from 0 to 1 changes the evidence for odd parity with the signs +, −, −, + for the histories 00, 01, 10 and 11, as parity requires (**Extended Data Fig. 3j; Methods**), and disabling the cubic term or resetting the state between segments breaks this sequence and reduces accuracy to 62.5% (**Extended Data Fig. 3k,l**). Without a phase cue, the trained nonlinear $R^2$NN therefore relies on both state retention and nonlinear mixing of the stored state with the current input. Circuit simulations of the mapped nonlinear network reproduce the mechanical probe responses, transients and cumulative energies for all eight patterns (**Extended Data Fig. 4; Methods; Supplementary Note 5**), and **Supplementary Note 6** analyzes its sensitivity to input, model and circuit parameters.

**2. Learned frequency selectivity of $R^2$NN**

Beyond recurrent memory, the $R^2$NN learns to route signals according to their spectral content. The voltage response of an output unit cell is

$$U_{\mathrm{out}} = \beta U_{in} = \beta Z_{\mathrm{eff}} I_{\mathrm{b}} = \frac{\omega_0^2}{\omega^2(\omega^2-\omega_1^2)D_M} I_{\mathrm{b}} = H(j\omega) I_{\mathrm{b}}, \tag{7}$$

where $I_{\mathrm{b}}$ is the current in the target branch, and $H(j\omega)$is a transresistance determined by the local parameters of the output cell. The output thus combines a local factor, reshaped by $R_{\mathrm{n}}$ through the local resonance, and a global factor, set by the coupling resistances $R_{\mathrm{c}}$ that distribute current among branches.

Training tunes both to form distinct impedance landscapes for inputs at different frequencies.

We illustrate this by training the metacircuit to classify noisy Hann-windowed pulses with carrier frequencies of 30, 50 and 70 Hz (**Fig. 3a; Supplementary Note 10**). Training reshapes local resonance frequencies throughout the lattice rather than only at the output cells, which serve as probes P1–P3 **(Fig. 3b)**. Between epochs 3 and 60, the largest changes occur along the column linking the input to P3, whose resonance rises from 56.1 to 77.8 Hz. This lowers the 70-Hz impedance of the P3 branch (**Fig. 3c**) and concentrates the initially diffuse 70-Hz current along a path to P3 (**Fig. 3d**); the 30- and 50-Hz currents are likewise channeled into P1 and P2. Because the output-cell resonances (33.5, 57.4 and 77.8 Hz) do not simply match the target carriers, frequency selectivity emerges from the combined tuning of local resonances and global coupling (**Supplementary Note 11**).

We implement the trained network on a separate 5 × 5 metacircuit (**Methods**). The measured transfer functions of P1, P2 and P3 peak near 30, 50 and 70 Hz, each within the band of its target class (**Fig. 3e**). Peak frequencies and line shapes agree with theory, whereas peak amplitudes differ mainly owing to deviations in resonator quality factor (**Supplementary Note 12**). The P3 peak lies near 70 Hz, below the 77.8-Hz resonance of its output cell, consistent with the peak being set by network-wide current routing. For each class, the target probe dominates the response within the 1-s integration window, at the end of which the corresponding output is set high (**Fig. 3f**), and all 60 unseen noisy test samples (20 per class) are classified correctly (100%; 95% Wilson interval, 94.0–100%; **Supplementary Note 10**). The frequency selectivity of the $R^2$NN is thus learned by the network as a whole, through the training-induced tuning of both local resonances and global current routing.

**Real-time analog temporal inference across physical domains.**

We next deploy integrated $R^2$NN systems on tactile, acoustic and vibration signals, reusing the pulse network for tactile perception and training new networks for the other tasks. Because the classes differ mainly in spectral content, these demonstrations primarily exploit the learned frequency selectivity.

**1. Frequency-coded tactile perception.**

We demonstrate a proof-of-concept tactile perception framework in which a robotic prosthesis equipped with the $R^2$NN recognizes frequency-coded textures representing Braille symbols (**Fig. 4a**). Each six-dot symbol is converted into three textured segments, one per row of the 3 × 2 dot matrix;

during sliding, these generate signals centered at 30, 50 and 70 Hz that encode "01", "10" and "11", whereas flat "00" segments produce no output. The $R^2$NN classifies each segment by its frequency, and the symbol is recovered from the sequence of codes. The conditioned signal of a fingertip pressure sensor sliding across the textures is fed to the battery-powered $R^2$NN system, whose logic outputs drive two vibration motors that convey the two bits of each code to the user (**Fig. 4b**). For the sequence "USTC", the $R^2$NN channels each segment into the output assigned to its frequency while the other outputs remain weak (**Fig. 4c**), correctly classifying all segments and decoding all four characters (**Fig. 4d** for "T", which contains all three textures; **Supplementary Video 1**). These results demonstrate real-time recognition of frequency-coded tactile textures with the $R^2$NN hardware and point to applications in assistive sensing and human–machine interaction. Combined with high-resolution tactile sensors[38], the approach could be extended to more compact textured surfaces and to conventional raised-dot Braille reading.

**2. Voice-command classification.**

Conventional speech pipelines digitize the waveform at audio rates and extract spectral features before classification[39], whereas in the $R^2$NN the continuous waveform is processed directly by the trained metacircuit. We train an $R^2$NN to classify three commands from a single speaker, "Hou" (backward), "Qian" (forward) and "Ting" (stop), assigned to P1, P2 and P3, and use it to control a humanoid robot (**Fig. 5a**). The dataset contains 100 separately recorded utterances per command without augmentation, 80 for training and 20 for testing (**Supplementary Note 10**). The working peaks of the measured transfer functions, defined as those carrying most of the energy of the target command, lie at 189, 108 and 120 Hz, close to the dominant spectral components of the corresponding commands at 185, 111 and 120 Hz (**Fig. 5b; Supplementary Note 12**). The system classifies 52 of 60 test recordings correctly (86.7%; 95% Wilson interval, 75.8–93.1%), with accuracies of 90%, 75% and 95% for "Hou", "Qian" and "Ting" (**Fig. 5c**). Most errors occur between "Qian" and "Ting", whose spectra overlap over 115–140 Hz. In representative recordings, the probe of the spoken command dominates the cumulative energy within the 1-s window, and the system output directly triggers the commanded robot motion in real time (**Fig. 5d,e; Supplementary Video 2**). Robustness to input amplitude, noise and frequency shift is characterized in **Supplementary Note 13**. Because online inference does not require buffering the speech waveform, the metacircuit could serve as a real-time, privacy-preserving front end for voice control in embedded systems.

**3. Condition monitoring of a quadrotor drone.**

Reliable operation of mechanical equipment requires continuous condition monitoring, for which vibration is a primary observable because it directly reflects the dynamic state of the system[7]. We demonstrate such continuous monitoring on a quadrotor drone with three operating states (**Fig. 6a**): normal operation, with a fundamental vibration frequency near 70 Hz; single-blade stall, which adds a component near 61 Hz; and four-blade stall, which lowers the fundamental to about 51 Hz. The working peaks of P1–P3 lie at 73, 60 and 51 Hz, each within the dominant band of its target state (**Fig. 6b; Supplementary Note 10**). On 180 held-out 1-s windows taken from the end of each state recording, the system classifies 178 correctly (98.9%; 95% Wilson interval, 96.0–99.7%; **Supplementary Note 14**). Scored against the true state over an entire 300-s run with four state transitions, the system output is correct 98.1% of the time (**Fig. 6c–e**); all errors fall within 2 s after a state change, and the status LEDs track the state in real time (**Fig. 6f; Supplementary Video 3**). These results show that the $R^2$NN can monitor the drone continuously and in real time, pointing to embedded diagnostics for industrial edge computing.

## Discussion

We have demonstrated an analog $R^2$NN in which a trained recurrent network is realized by the dynamics of a physical system: the hidden state is the oscillation of coupled local resonators, the recurrent weights are encoded in the admittance of the network and the state update is the continuous-time response of the metacircuit to its input. This is made possible by a reformulated MEA, which maps a trained locally resonant network one-to-one onto a metacircuit in which every trained parameter is a resistance. Training tunes the local resonances together with the resistive couplings: responses to past inputs persist in the coupled resonators and interfere with later ones, providing recurrent memory, and the resulting impedance landscape routes each signal to its designated output. Because both mechanisms are set by resonance rather than by wavelength, a compact circuit processes signals in the frequency band of human motion, speech and machine vibration. Tracing each result to the underlying dynamics further shows that such hardware can be designed as a dynamical system rather than tuned as a black box. Local resonance may thus serve as a general building block for physical neural hardware that computes on temporal signals where they are sensed.

Among experimental PNNs (**Supplementary Table 15**), memristor crossbars[15,17], photonic[19] and phononic[18] chips and self-learning transistor networks[14] have mostly implemented feedforward mappings of static or visual data, and temporal visual processing has relied on the short-term memory of volatile devices combined with a trained readout[16]. Memristor-based[40,41] and spintronic[42] reservoirs, coupled spin-torque nano-oscillators[43] and other physical computing systems[1] have processed speech and other temporal signals, but typically after digital pre-encoding, such as cochlear filtering, masking or conversion into spike trains, and with untrained physical dynamics interpreted by separately trained, often offline, readouts. Among these works, the $R^2$NN is the only one that combines trained recurrent physical dynamics, transferred one-to-one to a metacircuit, with raw-signal input, a parameter-free readout and continuous processing from sensor to decision. Compared with acoustic or mechanical PNNs, the metacircuit also avoids the wavelength-to-size constraint at low frequencies and reduces the simulation-to-reality gap caused by fabrication variability, and its logic-level outputs integrate directly into electromechanical systems.

In the present prototype, measured peak amplitudes and quality factors deviate from theory, mainly because of the finite gain–bandwidth product (GBW) of the operational amplifiers and parasitics. The GBW also limits the operating band. Because all characteristic frequencies of the trained network scale with the sampling frequency used in parameter mapping, faster amplifiers or integrated GICs would raise this band, and in the MHz range the local resonators could be implemented as passive LC resonators. The system power is mostly static, whereas each inference adds a dynamic power of only 10–14 mW. Replacing the general-purpose amplifiers with low-power ones therefore reduces the $R^2$NN-board power 5.9-fold and the total system power about threefold at the same accuracy. Integrated amplifiers designed for this application could reduce the static power further (Supplementary Note 9). The nearest-neighbor topology grows linearly in component count and power (Supplementary Note 9), and a 7 × 7 network trains as reliably as the fabricated 5 × 5 network (Supplementary Note 6), but larger hardware arrays and topologies for more classes remain to be demonstrated. All networks were evaluated on held-out test sets, and the hardware networks tolerate wide ranges of input amplitude and noise, whereas larger frequency drifts lead to misclassification (Supplementary Notes 7, 13 and 14). Extending the training data to multiple speakers, diverse acoustic environments and sensor placements, and rejecting words outside the command set by exploiting the temporal structure of speech, are natural next steps.

Nonlinear circuit elements would bring the nonlinear $R^2$NN studied here into hardware. Because the metacircuit runs without reset, its intrinsic decay sets both how long it retains past input and how soon it is ready for the next one, and trainable damping via variable capacitive elements would make this time scale tunable, providing short- and long-term memory[44] matched to the rate at which inputs arrive. Actively tunable resistors would replace offline training with in situ learning[14] and task reconfiguration, and, combined with microelectronic integration and energy harvesting, such networks could support autonomous long-term edge deployment. More broadly, our results establish engineered resonant dynamics as a physical substrate for recurrent neural computation, opening a direction in which machine intelligence for real-world temporal signals is built from physical dynamics.

# Methods

## Unit-cell characterization

The fabricated unit cell has $D_M = 1.307 \times 10^{-11}$ Ω·F$^2$, $D_m = 3.530 \times 10^{-11}$ Ω·F$^2$ and $R_n = 1$ MΩ, which give $\omega_0/2\pi = 26.8$ Hz and $\omega_1/2\pi = 51.5$ Hz. It is driven by a voltage sweep $U_{sweep}(t)$ from 1 to 100 Hz over 200 s, generated by a waveform generator (SDG2122X, SIGLENT Technologies), and the excitation and responses are acquired synchronously with a data acquisition system (NI 9205, National Instruments). The short-time Fourier transform (STFT) is applied to all signals, and the amplitude at each frequency is extracted along the ridge of the time–frequency spectrum that follows the instantaneous input frequency, which suppresses residual responses excited earlier in the sweep. The amplification factor is $\beta = U_{out}/U_{in}$, where $U_{out}$ is the voltage across $D_m$ **(Extended Data Fig. 1a)**. To measure the effective impedance, a resistor $R_s = 1$ MΩ is connected in series with the unit cell **(Extended Data Fig. 1b)**. With $U_{in}$ the voltage across the series combination and $U_{cell}$ that across the unit cell,

$$Z_{eff}(j\omega) = \frac{U_{cell}(j\omega)}{U_{in}(j\omega) - U_{cell}(j\omega)} R_s, \tag{8}$$

and $D_{eff}$ follows from equation (3). For ideal lossless elements, $D_{eff}$, $Z_{eff}$ and $\beta$ are real. Their measured real and imaginary parts are compared with equations (2) and (3), evaluated with the element values above, on linear and symmetric-logarithmic axes in **Extended Data Fig. 1c,d**, and the residuals of the real parts are shown in **Extended Data Fig. 1e,f**.

## Recurrent formulation of the mechanical and metacircuit-based R$^2$NNs

The mechanical R$^2$NN is a lattice of locally resonant unit cells, each consisting of a primary mass $M$ and an internal mass $m$ connected by a spring of stiffness $k_n$; the primary masses of neighboring cells are coupled by springs of stiffness $k_c$. Collecting the displacements of all masses in $\mathbf{x}(t)$, the lattice obeys

$$\mathbf{M}\ddot{\mathbf{x}} + \mathbf{C}\dot{\mathbf{x}} + \mathbf{K}\mathbf{x} = \mathbf{f}(t), \tag{9}$$

where **M**, **C** and **K** are the mass, damping and stiffness matrices and $\mathbf{f}(t)$ is the external force, applied at the input mass. **K** is assembled from $k_n$ and $k_c$ and is the only matrix that contains trainable parameters. The four corner unit cells are fixed (homogeneous Dirichlet boundary condition), which corresponds to grounding their nodes in the metacircuit, and their degrees of freedom are eliminated from all matrices. Central-difference discretization with time step $\Delta t$ gives

$$\begin{bmatrix}\mathbf{x}_{t+1}\\ \mathbf{x}_t\end{bmatrix} = \begin{bmatrix}(\mathbf{I}+\boldsymbol{\Omega})^{-1}(2\mathbf{I}-\Delta t^2\mathbf{M}^{-1}\mathbf{K}) & (\mathbf{I}+\boldsymbol{\Omega})^{-1}(-\mathbf{I}+\boldsymbol{\Omega})\\ \mathbf{I} & \mathbf{0}\end{bmatrix}\begin{bmatrix}\mathbf{x}_t\\ \mathbf{x}_{t-1}\end{bmatrix} + \begin{bmatrix}\Delta t^2(\mathbf{I}+\boldsymbol{\Omega})^{-1}\mathbf{M}^{-1}\\ \mathbf{0}\end{bmatrix}\mathbf{f}_t, \quad \boldsymbol{\Omega} = \frac{1}{2}\Delta t\mathbf{M}^{-1}\mathbf{C} \tag{10}$$

which has the form of the RNN update $\mathbf{h}_t = \mathbf{W}^{\mathrm{h}}(\mathbf{K})\mathbf{h}_{t-1} + \mathbf{W}^{\mathrm{f}}\mathbf{f}_t$, with $\mathbf{h}_t = [\mathbf{x}_{t+1}; \mathbf{x}_t]$. The trainable stiffnesses enter only through the state-transition matrix $\mathbf{W}^{\mathrm{h}}$, whereas the input matrix $\mathbf{W}^{\mathrm{f}}$ is fixed by $\mathbf{M}$, $\mathbf{C}$ and $\Delta t$.

In the nonlinear R$^2$NN, the internal spring of each unit cell is a Duffing spring with restoring force $k_{\mathrm{n}}q + k_{\mathrm{cubic}}q^3$, where $q$ is the displacement of the internal mass relative to its primary mass. The cubic terms add a state-dependent force vector $\mathbf{g}(\mathbf{x})$ to the left-hand side of equation (9), and the update becomes

$$\mathbf{h}_t = \mathbf{W}^{\mathrm{h}}(\mathbf{K})\mathbf{h}_{t-1} + \mathbf{W}^{f}[\mathbf{f}_t - \mathbf{g}(\mathbf{x}_t)], \tag{11}$$

where $\mathbf{x}_t$ is the first block of $\mathbf{h}_{t-1}$, so that the nonlinearity acts on the stored state.

Under the reformulated MEA, $\mathbf{x}$, $\mathbf{f}$, $\mathbf{M}$, $\mathbf{C}$ and $\mathbf{K}$ are replaced by the node-voltage vector $\mathbf{u}$, the injected-current vector $\mathbf{i}$, the FDNR matrix $\mathbf{D}$, the capacitance matrix $\mathbf{C}_{\mathrm{e}}$ and the admittance matrix $\mathbf{Y}$. Each of the $N$ unfixed unit cells $j$ has a primary node and an internal node, which are grounded through $D_{\mathrm{M}}$ and $D_{\mathrm{m}}$, respectively, and connected to each other through the internal resistor $R_{\mathrm{n},j}$; the primary nodes of neighboring cells are connected through the coupling resistors $R_{\mathrm{c},jk}$. With $\mathbf{u} = [\mathbf{u}^{\mathrm{M}}; \mathbf{u}^{\mathrm{m}}]$,

$$\mathbf{D} = \begin{bmatrix}D_M\mathbf{I}_N & \mathbf{0}\\ \mathbf{0} & D_m\mathbf{I}_N\end{bmatrix}, \qquad \mathbf{Y} = \begin{bmatrix}\mathbf{L}_c + \mathbf{G}_n & -\mathbf{G}_n\\ -\mathbf{G}_n & \mathbf{G}_n\end{bmatrix}, \tag{12}$$

where $\mathbf{G}_n = \mathrm{diag}\left(R_{n,1}^{-1}, \dots, R_{n,N}^{-1}\right)$ and $\mathbf{L}_c$ is the weighted Laplacian of the coupling network, $(\mathbf{L}_{\mathrm{c}})_{jj} = \Sigma_{k\in\mathcal{N}(j)}R_{\mathrm{c},jk}^{-1}$ and $(\mathbf{L}_{\mathrm{c}})_{jk} = -R_{\mathrm{c},jk}^{-1}$ for $k \in \mathcal{N}(j)$. Here $\mathcal{N}(j)$ is the set of nearest neighbors of cell $j$, including the grounded corner cells, whose coupling resistors therefore contribute only to the diagonal of $\mathbf{Y}$. Each off-diagonal entry of $\mathbf{Y}$ corresponds to one resistor, and $\mathbf{Y}$ is symmetric and positive definite. Damping is represented by a capacitance $C_{\mathrm{g}}$ from every node to ground and a capacitance $C_{\mathrm{n}}$ across every internal resistor, the counterparts of the mechanical dampers $c_{\mathrm{g}}$ and $c_{\mathrm{n}}$ **(Supplementary Note 3)**, so that

$$\mathbf{C}_{\mathrm{e}} = C_{\mathrm{g}}\mathbf{I}_{2N} + C_{\mathrm{n}}\begin{bmatrix}\mathbf{I}_N & -\mathbf{I}_N\\ -\mathbf{I}_N & \mathbf{I}_N\end{bmatrix}, \tag{13}$$

the mechanical damping matrix $\mathbf{C}$ has the same structure, with $c_{\mathrm{g}}$ and $c_{\mathrm{n}}$. Substituting these matrices into equation (10) gives the recurrence of the metacircuit, which reduces to equation (4) for $\mathbf{C}_{\mathrm{e}} = \mathbf{0}$, with $\mathbf{W}^{\mathrm{h}}(\mathbf{Y})$ and $\mathbf{W}^{\mathrm{i}}$ as in equation (5).

### Training of the R$^2$NNs

Each R$^2$NN is trained as its mechanical counterpart in a digital framework, and the trained parameters are then transferred to circuit elements (see 'Parameter mapping and scaling'). Training is carried out in dimensionless time $\tau = t/\Delta t$, so that the time step is unity. The masses $M$ and $m$, the damping coefficients and, in the nonlinear R$^2$NN, $k_{\mathrm{cubic}}$ are fixed; only $k_{\mathrm{n}}$ and $k_{\mathrm{c}}$ are trained, corresponding to the trainable resistances of the metacircuit. Each training sample is applied as a force sequence at the input mass and propagated through the linear or nonlinear recurrence above. The probe energies $E_{\mathrm{prb},l}$, defined as in the main text, are computed from the displacements of the internal masses of the output cells over the whole sequence, and the class probabilities are obtained with $\log E_{\mathrm{prb},l}$ as logits,

$$p_\ell = \mathrm{softmax}\big(\log \mathbf{E}_{\mathrm{prb}}\big)_\ell = \frac{E_{\mathrm{prb},\ell}}{\sum_l E_{\mathrm{prb},l}}, \tag{14}$$

so that training optimizes the same normalized energies that are compared at inference. The loss $\mathscr{L} = \mathscr{L}_{\mathrm{CE}} + \lambda\mathscr{L}_{\mathrm{margin}}$ combines the cross-entropy between $p_l$ and the class labels with an energy-margin term, weighted by $\lambda$, that enlarges the energy gap between the target probe and the strongest competing probe. Gradients with respect to $k_{\mathrm{n}}$ and $k_{\mathrm{c}}$ are computed by backpropagation through time, and the parameters are updated with the Adam optimizer. Datasets, initialization, damping coefficients and hyperparameters for the individual tasks are given in Supplementary Notes 5 (temporal parity) and 10 (other tasks).

### Parameter mapping and scaling

We illustrate the mapping for a single Duffing oscillator; the same relations apply element by element to the network, with the displacement replaced by the relative displacement across each spring. Let $\Delta t$ be the physical time step and $f_{\mathrm{s}} = 1/\Delta t$ the sampling frequency. The trained model obeys $mx'' + \hat{c}x' + \hat{k}x + \hat{k}_{\mathrm{cubic}}x^3 = \hat{f}$, where primes denote differentiation with respect to $\tau$ and the hatted quantities are the normalized damping, linear stiffness, cubic stiffness and force. Because $\mathrm{d}/\mathrm{d}t = f_{\mathrm{s}}\,\mathrm{d}/\mathrm{d}\tau$, this is equivalent in physical time to

$$m\ddot{x} + \hat{c}f_s\dot{x} + \hat{k}f_s^2x + \hat{k}_{\mathrm{cubic}}f_s^2x^3 = f_s^2\hat{f}. \tag{15}$$

that is, the physical parameters are $c = \hat{c} f_s$, $k = \hat{k} f_s^2$, $k_{\text{cubic}} = \hat{k}_{\text{cubic}} f_s^2$, $f = \hat{f} f_s^2$. The sampling frequency thus sets the physical time scale of the network, and all of its characteristic frequencies scale in proportion to $f_s$. We then introduce a voltage scale $\gamma$, with $u = \gamma x$, and an element scale $\alpha$. Substituting $x = u/\gamma$ into this equation and multiplying by $\alpha\gamma$ gives the metacircuit form $D\ddot{u} + C\dot{u} + u/R(u) = i$, with

$$D = \alpha m,\ C = \alpha \hat{c} f_s,\ R(u)^{-1} = \alpha \hat{k} f_s^2 + \frac{\alpha \hat{k}_{\text{cubic}} f_s^2}{\gamma^2} u^2,\ i = \alpha\gamma f_s^2 \hat{f}. \qquad (16)$$

The scale $\alpha$ multiplies $D$, $C$ and $R^{-1}$ by the same factor and therefore sets the impedance level of the circuit without altering its dynamics; it is chosen to bring the element values into realizable ranges. The scale $\gamma$ sets the voltage swing. In linear networks it only rescales the input current and is chosen so that all node voltages remain within the linear range of the operational amplifiers; in nonlinear networks it also fixes the cubic coefficient of $R(u)$, so the operating amplitude and the nonlinear element must be set consistently. Neither factor changes the normalized dynamics learned in training. Linear $R^2$NNs are implemented in hardware, whereas the nonlinear element $R(u)$ is implemented in circuit simulation (LTspice). The scale factors and element values for the individual tasks are listed in Supplementary Notes 10 and 12.

**Metacircuit-based $R^2$NN system**

The stand-alone system combines the network with an input conditioning stage, a voltage-controlled current source (VCCS), an output conditioning stage and a microcontroller unit (MCU; STM32F103VET6, STMicroelectronics) **(Extended Data Fig. 2a)**. Each $R^2$NN is a 5 × 5 array of unit cells with the four corner cells grounded **(Extended Data Fig. 2b)**. The sensor voltage is buffered, a.c.-coupled and amplified with an adjustable gain, and the VCCS, an improved Howland current source with a transconductance of $g_m = 1$ μS, injects the corresponding current at the input node. Each class is assigned one output unit cell (three for the three-class tasks and two for the temporal-parity task), whose output is the voltage $U_{out}$ across its $D_m$ element. The output voltages are buffered and shifted by $V_{MID}$ into the input range of the analog-to-digital converter of the MCU. The MCU continuously samples a monitoring copy of the input; when the input exceeds a threshold, it accumulates the squared deviation of each output from its idle level over a fixed window, assigns the class with the largest energy and indicates the result on an LED or issues it as a device-control command. After each decision, the MCU clears its accumulators and returns to monitoring; in the drone task, where the vibration is continuous, it accumulates over consecutive 1-s windows without a gap. The metacircuit itself is not reset: it remains powered throughout each acquisition, and its state decays between inputs through the intrinsic losses of the circuit (**Supplementary Note 15**). All hardware tests were recorded in this mode, with the inputs following each other in continuous recordings. The system consists of the $R^2$NN board, an auxiliary conditioning board and an MCU board **(Extended Data Fig. 2c)** and is powered by a single 5-V battery. Circuit schematics are given in **Supplementary Note 4**.

### Frequency response of the trained $R^2$NNs

The trained networks are characterized with the same sweep method, with the sweep voltage converted into a current $I(t) = g_m U_{sweep}(t)$ by the VCCS and injected at the input node. The voltages $U_l(t)$ of the output channels are sampled at 10 kHz, and the transfer function of each output channel is obtained as

$$|T_l(j\omega)| = \frac{|U_l(j\omega)|}{|I(j\omega)|} = \frac{|U_l(j\omega)|}{g_\mathrm{m}|U_\mathrm{sweep}(j\omega)|}, \tag{17}$$

with $|U_l(j\omega)|$ extracted along the sweep ridge. The sweeps cover 1–100 Hz over 100 s for the temporal-parity, pulse and drone networks and 1–300 Hz over 300 s for the voice-command network.

### Temporal-parity task and mechanism analysis

**Dataset.** Of the three-bit sequences $\mathbf{b} = (b_1, b_2, b_3)$, patterns 001, 010, 100 and 111 form the odd-parity class and the remaining patterns the even-parity class **(Extended Data Fig. 3a)**. Each bit is a sinusoidal segment of duration $T = 0.2$ s sampled at $f_s = 1$ kHz, giving 600 samples per waveform (**Extended Data Fig. 3b**). For segment $j$ and sample $r$ within it, the input is

$$s_{j,r} = \sin\left[\phi_0 + 2\pi\left(T\sum_{i<j} f_i + \frac{r f_j}{f_\mathrm{s}}\right)\right] + \sigma\xi_{j,r},\ \ f_j = 40\ \mathrm{Hz} + (f_\mathrm{H} - 40\ \mathrm{Hz})b_j, \tag{18}$$

where $f_H = 60$ Hz for phase-neutral and 62 Hz for phase-accumulating encoding, $\phi_0$ is the initial phase of the sequence, drawn independently and uniformly from $[0, 2\pi)$ for each waveform, $\xi_{j,r}$ are independent standard Gaussian variables and $\sigma = 0.15$. Because each segment starts at the phase reached at the end of the previous one, the waveform has no phase jump at segment boundaries. Each dataset contains 2,000 training and 800 independent test waveforms, balanced across the eight patterns.

**Models and readout.** The simulated $R^2$NNs are $7 \times 7$ lattices with the four corner cells fixed

(**Extended Data Fig. 3c**), with $k_{\mathrm{cubic}} = 0$ in the linear and $k_{\mathrm{cubic}} = 2 \times 10^{-10}$ in the nonlinear R$^2$NN, in the normalized units used for training. The class is read from the probe energies $E_{\mathrm{prb},1}$ and $E_{\mathrm{prb},2}$ of two output cells (degrees of freedom 70 and 76) through $\Gamma = E_{\mathrm{prb},2} - E_{\mathrm{prb},1}$, which is positive for odd and negative for even parity. Five-fold cross-validation is performed within the training pool, after which each final model is retrained on all 2,000 training waveforms and evaluated on the independent test set. Shading in **Extended Data Fig. 3d–f** denotes the mean ± 1 s.d. across folds.

**Phase controls for the linear R$^2$NN.** Both controls in **Extended Data Fig. 3h** use the linear checkpoint trained with phase-accumulating encoding, without retraining. In the phase-reset control, the accumulated-phase term of the input is removed, so that every segment starts from the same source phase while the network state remains continuous. In the encoding-transfer control, the checkpoint is applied to the 800 phase-neutral test waveforms.

**Mechanistic interventions on the nonlinear R$^2$NN.** All interventions are applied at inference to the fixed nonlinear checkpoint, without retraining. The cubic ablation sets only $k_{\mathrm{cubic}}$ to zero. The state-reset ablation starts each 200-sample segment from zero displacement and velocity while retaining the original waveform, nonlinear parameters and accumulated probe readout **(Extended Data Fig. 3l)**. To separate memory from nonlinear mixing, each test waveform is split into a history input containing only the first two segments and a current input containing only the third, and the full, history and current inputs are simulated separately, giving the probe displacements $x^{\mathrm{full}}$, $x^{\mathrm{hist}}$ and $x^{\mathrm{cur}}$. Residual memory and non-additivity are

$$\rho = \sqrt{\langle (x^{\mathrm{hist}})^2 \rangle}, \quad \eta = \sqrt{\frac{\langle (x^{\mathrm{full}} - x^{\mathrm{hist}} - x^{\mathrm{cur}})^2 \rangle}{\langle (x^{\mathrm{full}})^2 \rangle}}, \tag{19}$$

where $\langle \cdot \rangle$ denotes the mean over the third segment, both probes and all test waveforms; $\eta = 0$ for exact linear superposition (**Extended Data Fig. 3i**). For the history $b_1 b_2$, the effect of the third bit is $\Delta_3(b_1 b_2) = \overline{\Gamma}(b_1 b_2 1) - \overline{\Gamma}(b_1 b_2 0)$, where each mean is taken over the 100 test waveforms of a pattern; error bars in **Extended Data Fig. 3j** denote one standard error of the difference. **Extended Data Fig. 3k** shows the normalized evidence $\Gamma/(E_{\mathrm{prb},1} + E_{\mathrm{prb},2})$ as the mean ± s.e.m. over the 100 test waveforms of each pattern.

**Circuit simulation.** For **Extended Data Fig. 4**, one test waveform of each pattern is applied to the nonlinear mechanical model and to its LTspice implementation, and the two probe responses are compared after amplitude conversion.

**Hardware experiment.** The 5 × 5 linear $R^2$NN trained with phase-accumulating encoding is implemented as a metacircuit, with its two probes as output channels. For each of the eight patterns, 50 waveforms drawn from a newly generated independent test set are produced by the waveform generator, converted into currents by the VCCS and injected at the input node, giving 400 test waveforms. The two probe voltages are recorded at 10 kHz with the data acquisition system, and the probe energies are computed offline as the sums of the squared probe voltages over the 0.6-s sequence. In parallel, the MCU accumulates the squared probe voltages over the same window and assigns the class to the probe with the larger energy.

## Statistical analysis

Classification accuracies are reported as the number $n_{\text{corr}}$ of correctly classified samples out of the $n_{\text{test}}$ samples of a test set, with 95% Wilson score intervals $[\hat{a} + z^2/(2n_{\text{test}}) \pm z(\hat{a}(1 - \hat{a})/n_{\text{test}} + z^2/(4n_{\text{test}}^2))^{1/2}]/(1 + z^2/n_{\text{test}})$, where $\hat{a} = n_{\text{corr}}/n_{\text{test}}$ is the observed accuracy and $z = 1.96$. Unlike the normal approximation, this interval remains informative when all samples are classified correctly, in which case its lower bound is $n_{\text{test}}/(n_{\text{test}} + z^2)$. Cross-validation results are given as the mean ± sample standard deviation over the five folds. For the drone task, the test windows are held out in time from the same recording of each state, so the interval does not account for correlations between windows of one recording. Each test set was evaluated once, after the final model had been fixed, and no test data were used for model selection.

## Acknowledgements

This work was supported by the National Natural Science Foundation of China under Grants No. 12472094, No. 12532001, No. 52275116, No. 52105112; the Fundamental Research Funds for the Central Universities in China under Grants No. YD2090002019, No. WK2090000068; the Research Project of State Key Laboratory of Mechanical System and Vibration (Grant No. MSV202403); the Aeronautic Science Foundation of China (Grant No. 2023Z009078001); the Major Project of Anhui Province's Science and Technology Innovation Breakthrough Plan (202423h08050003).

## Author contributions

T.J. conceived the original idea and designed the project. Z.Z. and T.J. performed the theoretical analysis, model design, experimental measurements, and data analysis. Z.Z., M.Y., and T.J. demonstrated the applications of metacircuit-based $R^2$NNs. M.S. and Y.Z. assisted with the experimental analysis and the revision of the manuscript. Z.F. provided guidance on circuit design. T.J. and Z.Z. wrote the original manuscript. T.J., S.Z., and Q.H. supervised the work. All authors contributed to the analysis and discussion of the results and to the revision of the manuscript.

## Competing interests

The authors declare that they have no competing interests.

## Data availability

The authors declare that the main data supporting the findings of this study are available within the article and its Supplementary Information files. The datasets and raw sensor recordings generated in this study will be made publicly available upon publication. Hardware source data are provided as Supplementary Data 1, and other data are available from the corresponding author upon request.

## Code availability

The code for training the $R^2$NNs and the LTspice netlists of the metacircuits will be made publicly available upon publication. Other code used to generate the results in this paper and the Supplementary Information is available from the corresponding author upon request.

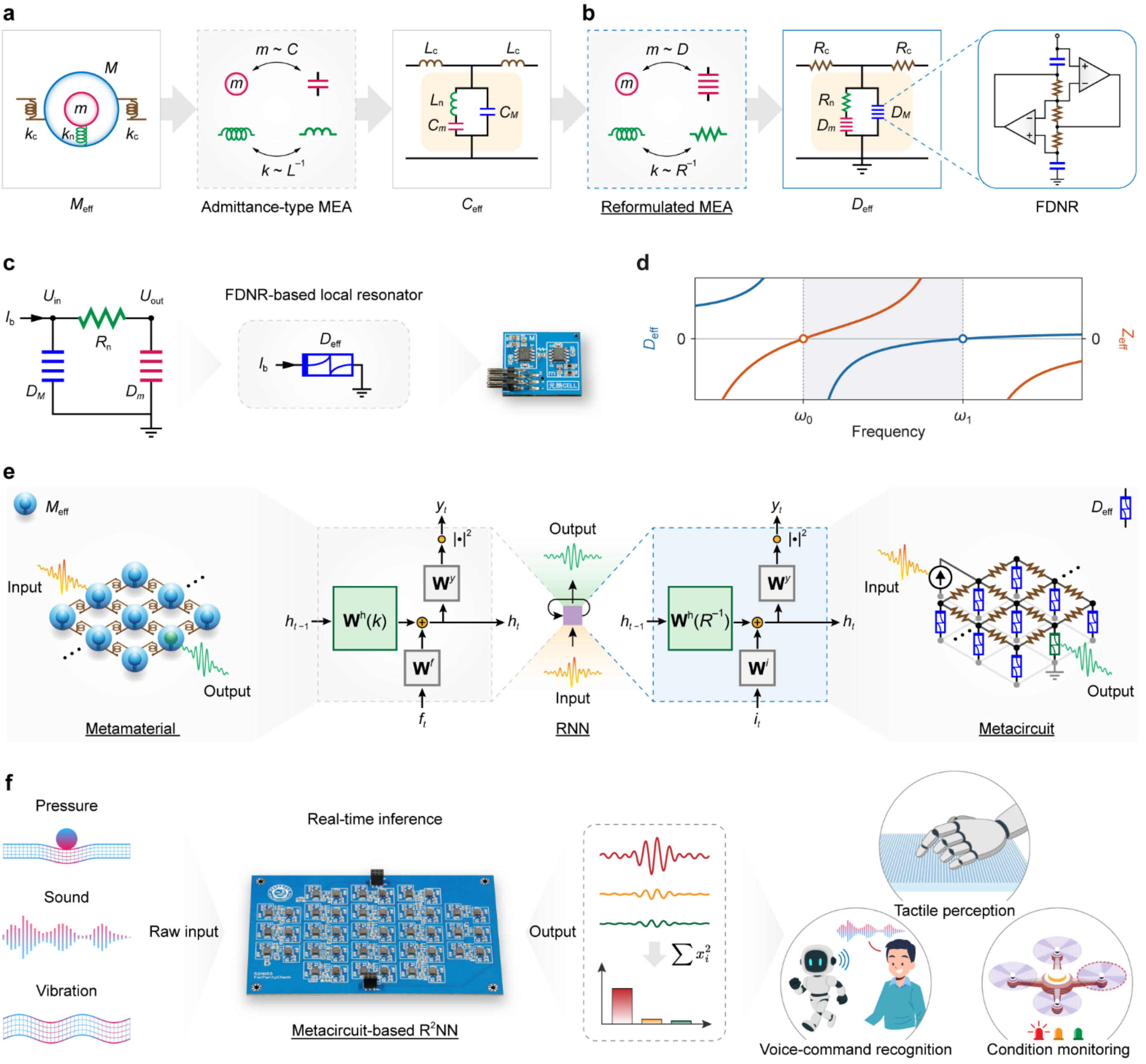

**Figure 1. Concept of the metacircuit-based R²NN. a**, Equivalence between a mechanical local resonator (masses $M$ and $m$, stiffnesses $k_n$ and $k_c$, effective mass $M_{eff}$) and an electrical LC local resonator (capacitances $C_M$ and $C_m$, inductances $L_n$ and $L_c$, effective capacitance $C_{eff}$) under the conventional admittance-type mechanical–electrical analogy (MEA). **b**, Reformulated MEA, yielding a physically realizable FDNR-based local resonator (effective FDNR element $D_{eff}$). Each FDNR is implemented with a generalized impedance converter (GIC). **c**, Circuit of the FDNR-based local resonator (left), its schematic symbol $D_{eff}$, which reflects its frequency characteristics (middle), and a photograph of the fabricated unit cell (right). **d**, Frequency dependence of $D_{eff}$ and $Z_{eff}$. **e**, Mapping among a mechanical metamaterial, an RNN, and the metacircuit. The state-transition weights $\mathbf{W}^h$ are determined by the spring stiffnesses $k$ in the metamaterial and by the conductances $R^{-1}$ in the metacircuit. **f**, The metacircuit-based R²NN performs direct analog inference on dynamic physical signals (pressure, sound and vibration) injected as currents. Classification relies on the accumulated output energy, compared by a readout without trainable parameters, enabling applications such as frequency-coded tactile perception, voice-command classification and drone condition monitoring.

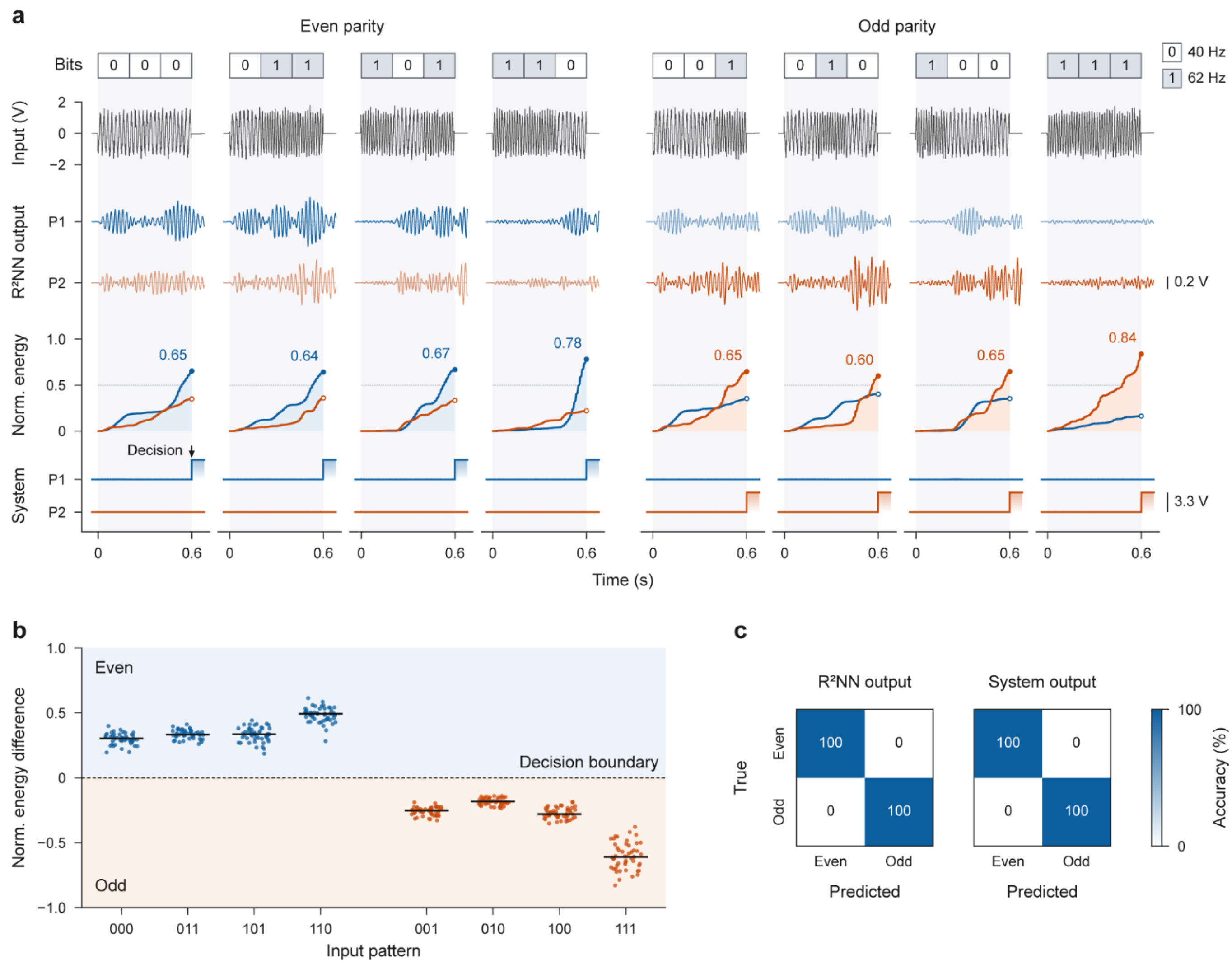

**Figure 2. Experimental temporal-parity classification by the R²NN. a**, Representative single-shot recordings for the eight patterns, grouped by parity: even (000, 011, 101, 110), assigned to probe P1, and odd (001, 010, 100, 111), assigned to probe P2. From top to bottom: bit code; amplified input voltage, in which each bit is a 0.2-s segment at 40 Hz (0) or 62 Hz (1); probe voltages at P1 and P2, with the non-target probe in a lighter shade (scale bar, 0.2 V); normalized cumulative energy of each probe, with the final value of the dominant probe labelled; and system outputs (logic level, 0–3.3 V). $t = 0$ marks the detection of the input, and grey bands mark the 0.6-s integration window. **b**, Normalized energy difference $(E_1 - E_2)/(E_1 + E_2)$ for all 400 recordings, where $E_i$ is the sum of the squared P$i$ voltage over the integration window. Black bars, medians; dashed line, decision boundary ($E_1 = E_2$). **c**, Confusion matrices for the same 400 recordings, with the probe energies compared offline from the recorded probe voltages (left) and in real time by the system (right). Both give 400/400 correct classifications.

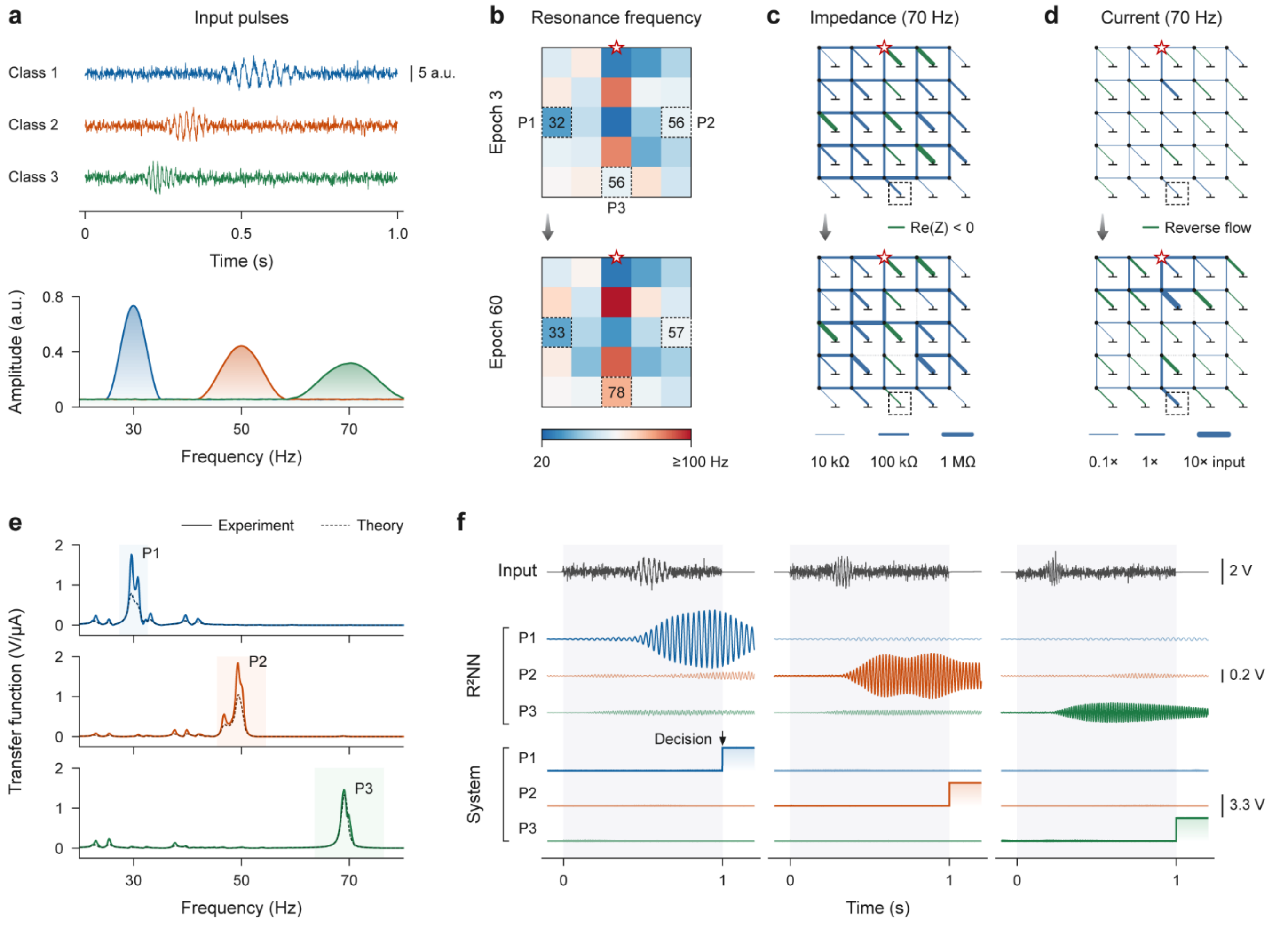

**Figure 3. Learned frequency selectivity of the R$^2$NN. a**, Representative input pulses of the three classes (top) and their amplitude spectra (bottom). **b**, Local resonance frequencies of the unit cells at training epochs 3 and 60. Star, input cell; dashed boxes, output cells (probes P1–P3), labelled with their resonance frequencies (Hz). **c**,**d**, Impedance (**c**) and current (**d**) distributions of the network at 70 Hz at epochs 3 and 60. Line width indicates the impedance magnitude in **c** and the branch current relative to the input current in **d**; green marks branches with Re($Z$) < 0 in **c** and reverse current flow in **d**. The dashed box marks the P3 branch. **e**, Measured (solid) and theoretical (dashed) transfer functions of P1–P3, offset vertically. **f**, Measured input, probe voltages of the R$^2$NN and system outputs for one representative sample of each class. Grey bands mark the 1-s integration window, at the end of which the system output switches.

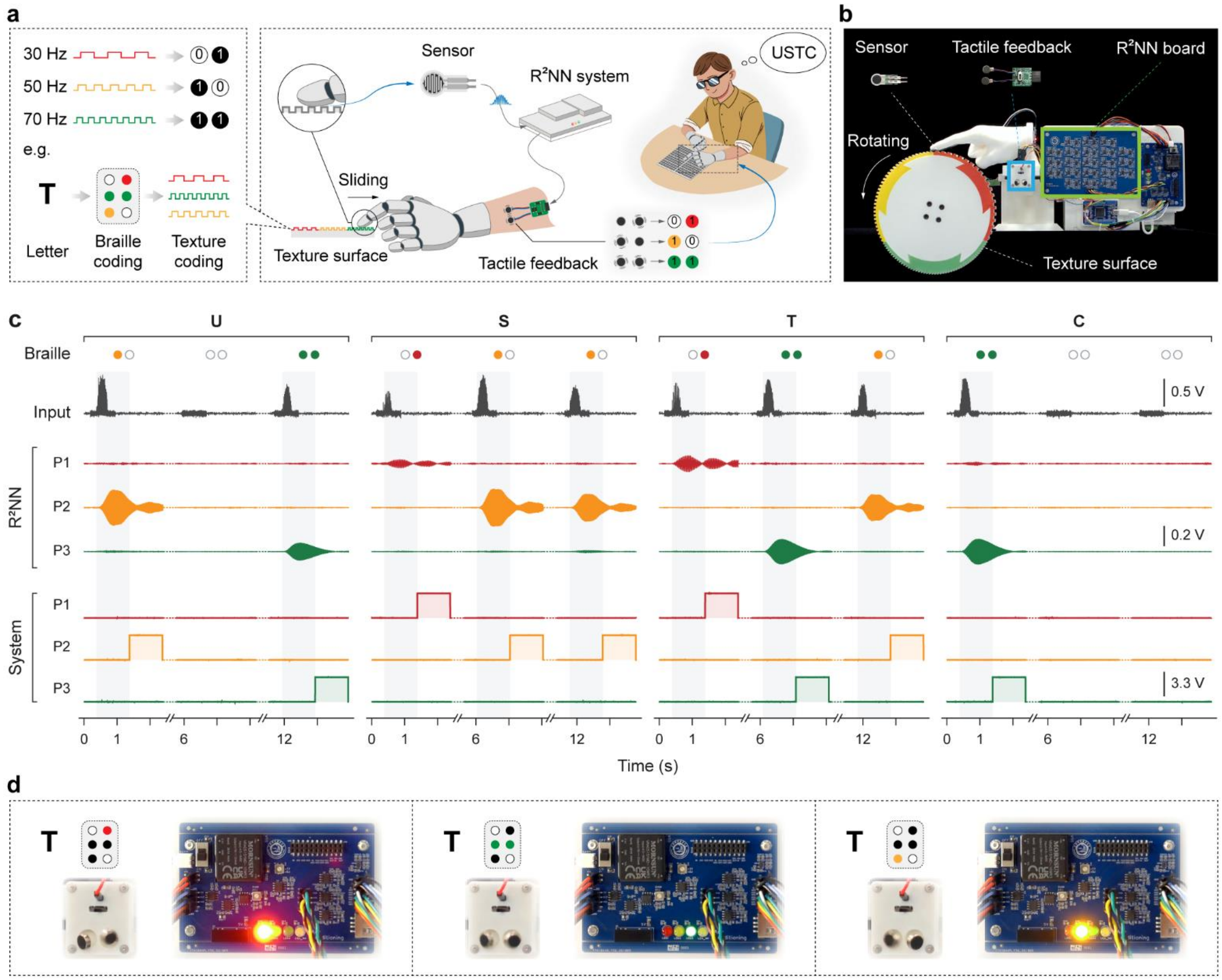

**Figure 4. Recognition of frequency-coded tactile textures with the R$^2$NN. a**, Concept and signal pipeline. A robotic prosthesis equipped with the R$^2$NN is envisioned as a tactile aid for users with visual and manual impairments. Each Braille character is encoded as three textured segments, one per row of the 3 × 2 dot matrix, with textures of 30, 50 and 70 Hz encoding the row codes “01”, “10” and “11”, respectively, and flat segments encoding “00” (shown for “T”). The signal from the fingertip sensor is processed by the R$^2$NN system, whose logic outputs drive a tactile feedback module with one vibration motor per bit. **b**, Experimental setup: simulated hand with a pressure sensor on the fingertip, rotating textured surface, R$^2$NN system and tactile feedback module. **c**, Tactile signals, probe voltages of the R$^2$NN and system outputs recorded during recognition of the characters “U”, “S”, “T” and “C”. Dots above each segment give the row code and are colored by the expected output. **d**, Tactile feedback for “T”: vibration cues generated for each of its three row codes and the corresponding output indicators (Supplementary Video 1).

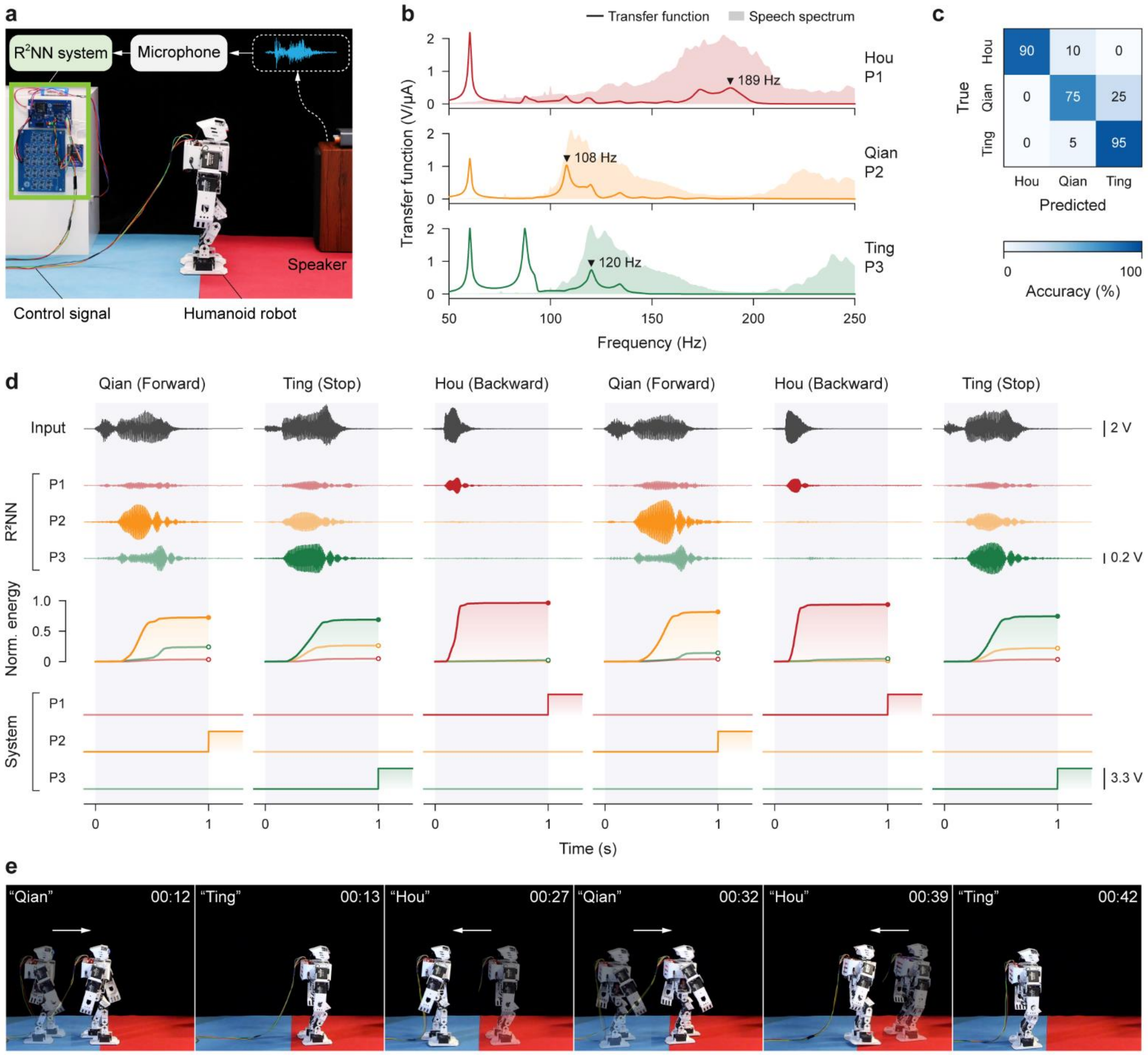

**Figure 5. Real-time voice-command classification and robotic control with the R²NN. a**, System schematic. A command spoken by a single speaker is captured by a MEMS microphone, conditioned and injected into the R²NN system without digitization, and the control board drives the humanoid robot from the system output. **b**, Average amplitude spectra of the training recordings of the three commands ("Hou"/backward, "Qian"/forward and "Ting"/stop; each normalized to its maximum between 50 and 250 Hz, arbitrary units; shaded) and measured transfer functions of the corresponding probes P1–P3 (lines). Triangles mark the working peak of each probe, that is, the peak carrying most of the energy of its target command. **c**, Confusion matrix of the R²NN classification for the 60 test recordings ($n$ = 20 per command; 52/60 correct). **d**, Six representative test recordings, two per command, in the order of the robot demonstration. From top to bottom: amplified input, probe voltages, cumulative energy of each probe, normalized to the sum of the final energies of all probes, and system outputs. **e**, Motions of the humanoid robot triggered by the system outputs in **d** (Supplementary Video 2).

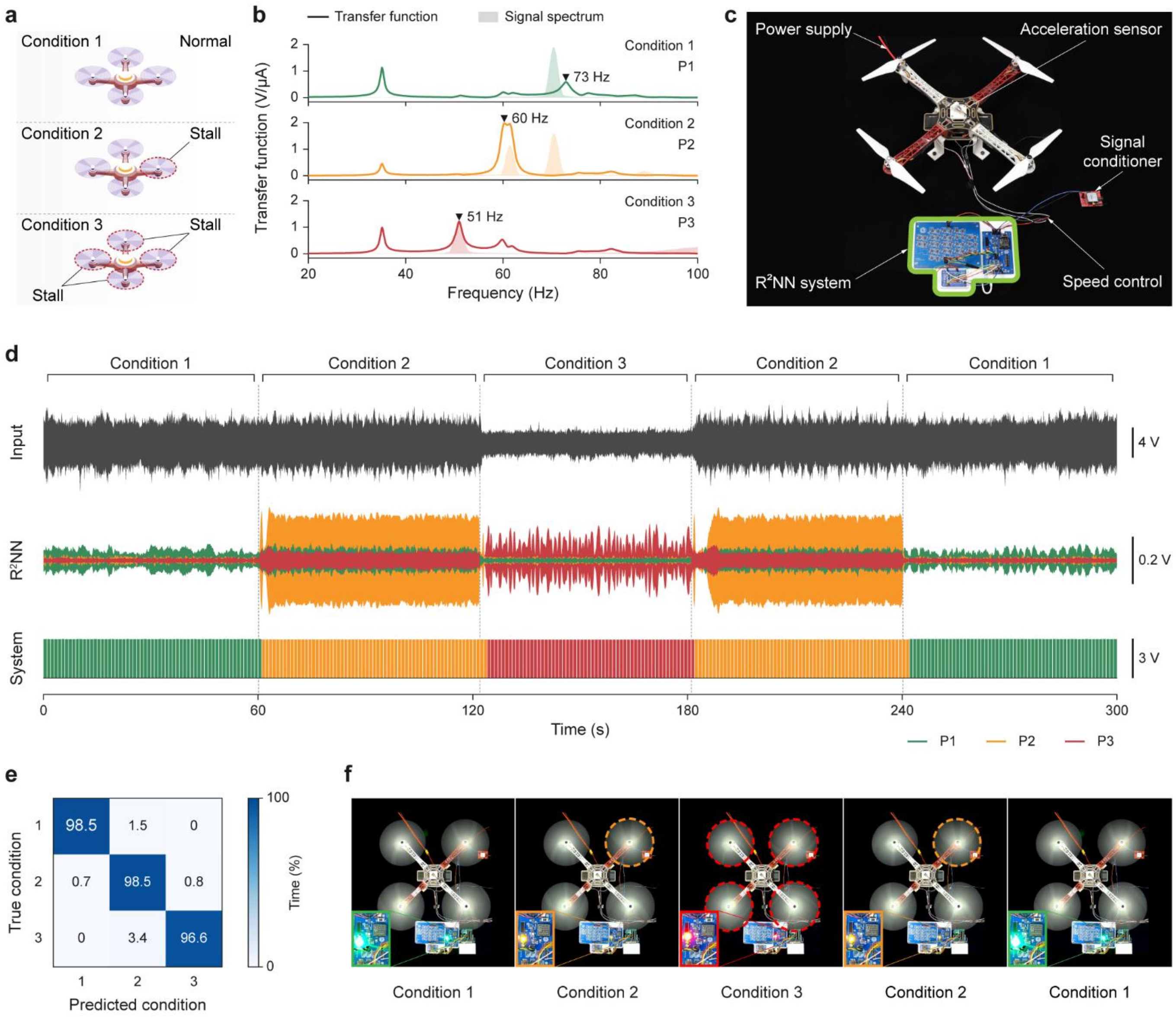

**Figure 6. Continuous condition monitoring of a quadrotor drone with the R$^2$NN. a**, The three operating states: normal operation (Condition 1), single-blade stall (Condition 2) and four-blade stall (Condition 3). **b**, Measured transfer functions of probes P1–P3 (lines) and normalized amplitude spectra (arbitrary units) of the vibration signals of the corresponding states. Triangles mark the working peak of each probe. **c**, Experimental setup used to verify real-time, long-duration drone condition monitoring. **d**, Continuous 300-s run with four state transitions: sensor voltage (input), probe voltages of the R$^2$NN and system outputs. Dashed lines mark the state transitions (Supplementary Note 14), and brackets give the true state. White lines on the system outputs mark the 1-s decision instants; the output level is held between decisions. **e**, Time-resolved confusion matrix of the system output against the true state over the entire run, including the transitions, given as a percentage of the duration of each true state. **f**, Status LEDs on the R$^2$NN system during the five phases of the run (Supplementary Video 3).

# Extended Data Figures

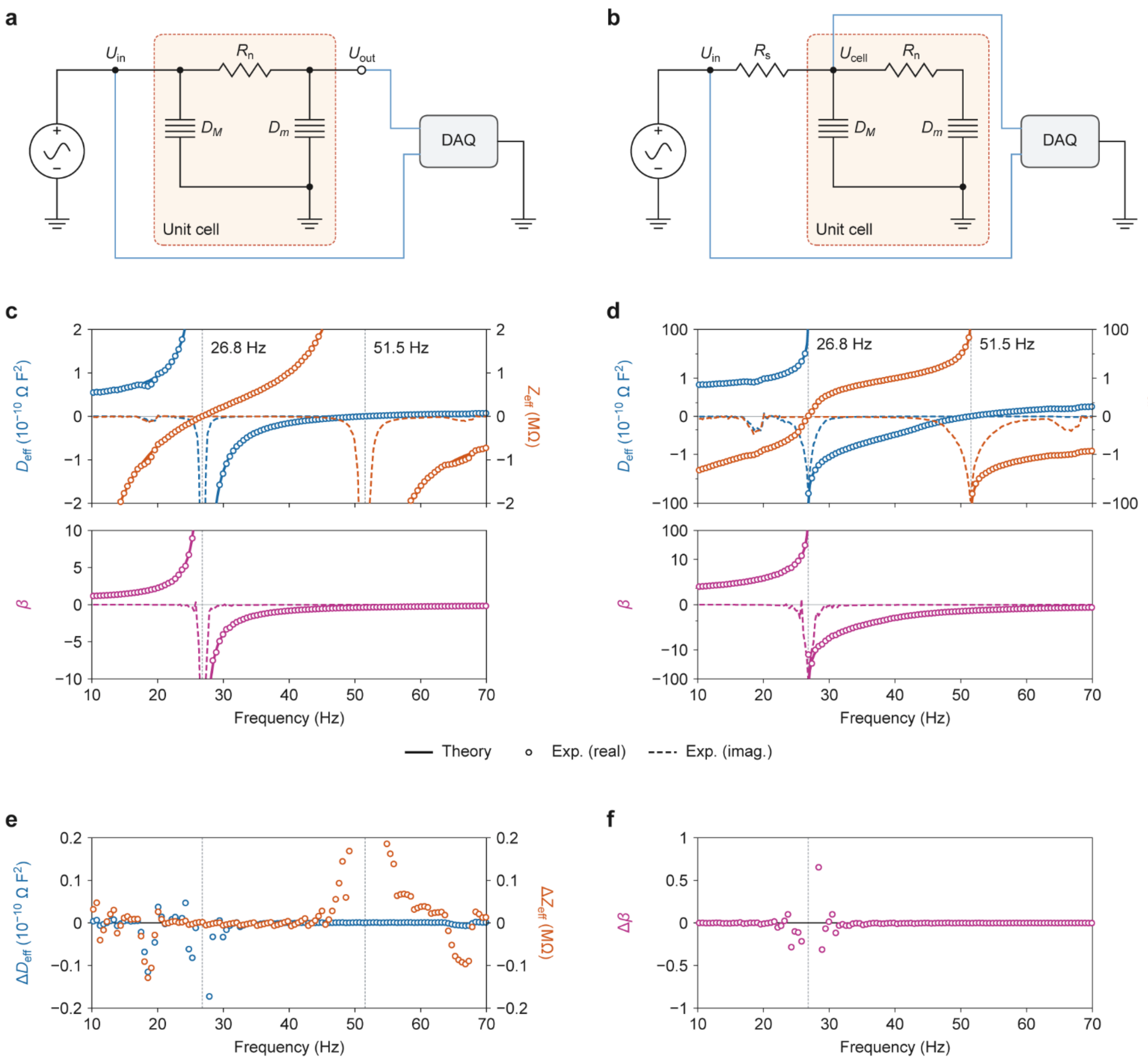

**Extended Data Figure 1. Characterization of the FDNR-based unit cell. a**,**b**, Circuits for measuring the amplification factor $\beta$ and the effective impedance $Z_{\mathrm{eff}}$. **c**,**d**, Measured $D_{\mathrm{eff}}$, $Z_{\mathrm{eff}}$, and $\beta$. **c** uses linear axes and **d** uses symmetric-logarithmic axes. Solid lines, theory; circles and dashed lines, measured real and imaginary parts, respectively. Vertical dashed lines mark $\omega_0/2\pi = 26.8$ Hz and $\omega_1/2\pi = 51.5$ Hz. **e**,**f**, Residuals of the real parts in **c** for $D_{\mathrm{eff}}$ and $Z_{\mathrm{eff}}$ (**e**) and $\beta$ (**f**).

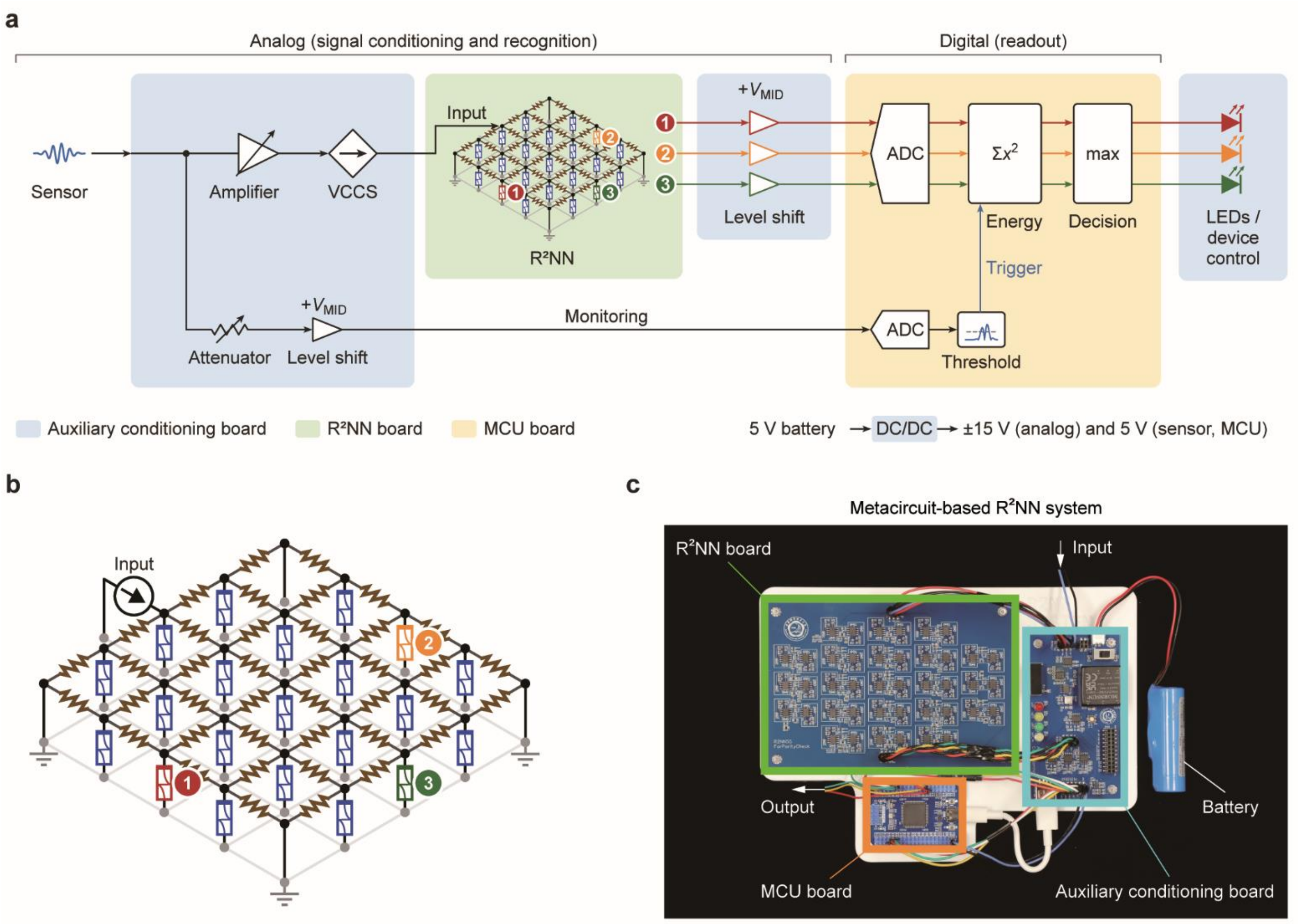

**Extended Data Figure 2. Metacircuit-based R$^2$NN system.** **a**, Signal chain. The sensor signal is buffered, a.c.-coupled, amplified and converted into a current by a voltage-controlled current source (VCCS), which drives the R$^2$NN at its input node. The voltages of the three output unit cells (1–3, one per class) are shifted by $V_{MID}$ into the input range of the analog-to-digital converter (ADC) of the MCU. A monitoring channel, sampled by the same ADC, triggers the energy accumulation when the input exceeds a threshold; the readout then selects the channel with the largest energy, and the result is shown on indicator LEDs or used for device control. Shaded regions indicate the circuit boards. **b**, Topology of the metacircuit-based R$^2$NN, a 5 × 5 array of unit cells with the four corner unit cells grounded. The current is applied at the input node; the red, orange and green unit cells are output unit cells 1–3, numbered by output channel as in **a**. **c**, Photograph of the system, consisting of the R$^2$NN board, the auxiliary conditioning board and the MCU board.

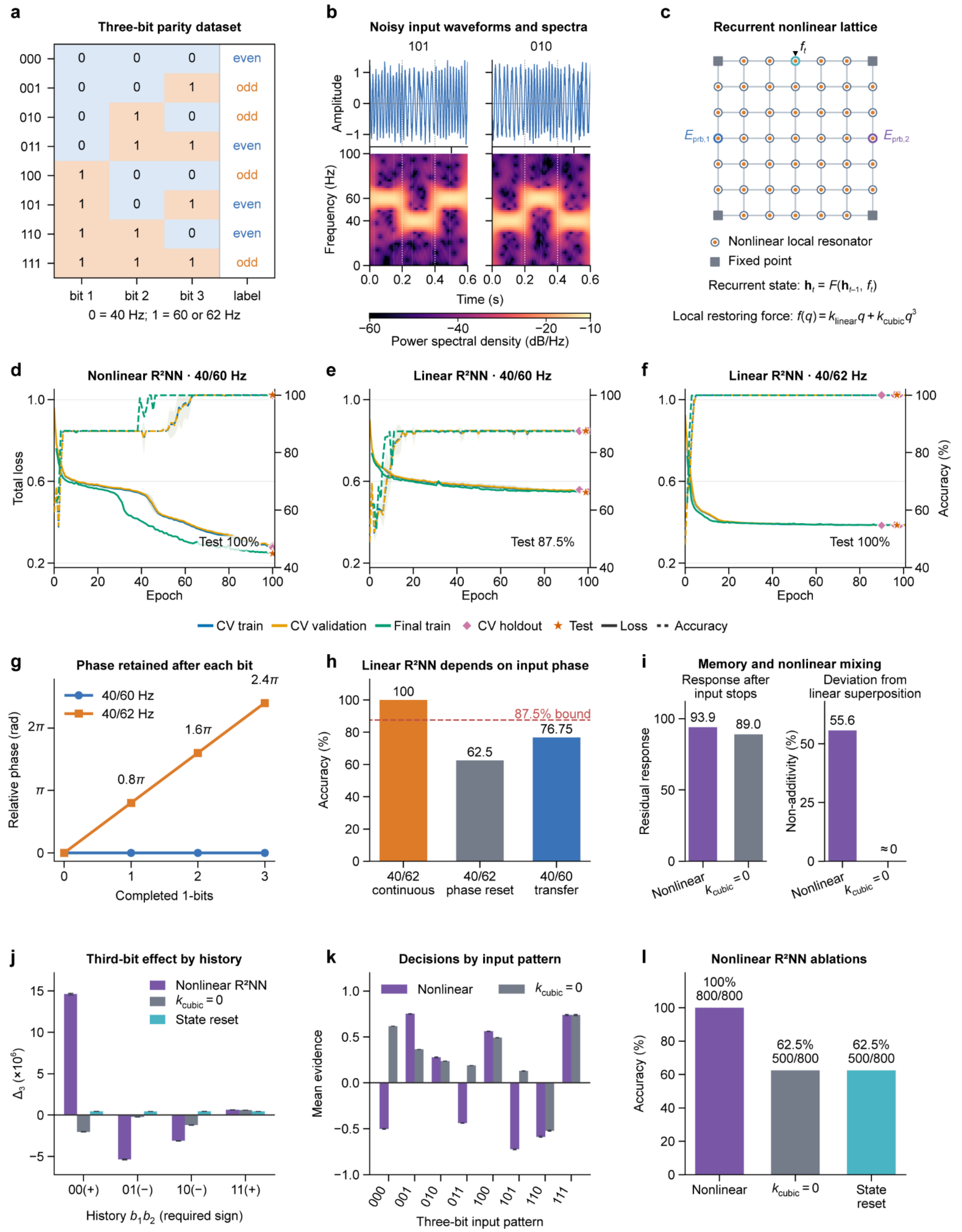

**Extended Data Figure 3. Recurrent and nonlinear mechanisms of the R²NN in the temporal-parity task.** **a**, Eight three-bit input patterns and their parity labels. **b**, Representative noisy waveforms and time-frequency maps under phase-neutral encoding; dotted lines mark segment boundaries. **c**, The 7 × 7 mechanical R²NN, showing the input site, readout probes, nonlinear local resonators and fixed nodes. **d**–**f**, Training trajectories

of the nonlinear model with phase-neutral encoding, the linear models with phase-neutral and phase-accumulating encoding, respectively. The cross-validation (CV) training and validation curves show the mean across five folds, with shading indicating ±1 standard deviation across folds. The final-training curves show retraining on the full training pool. Diamonds and stars mark CV holdout and independent-test results, respectively. Solid and dashed lines denote loss and accuracy. **g**, Relative phase accumulated as successive 1-bits are completed. **h**, Phase-reset and encoding-transfer controls for the linear checkpoint trained with phase-accumulating inputs. The dashed line marks the 87.5% ceiling reached after retraining the linear $R^2NN$ directly with phase-neutral inputs. **i**, Direct tests of memory and nonlinear response mixing across all 800 phase-neutral test waveforms. Left, residual probe response during the 0.2 s after input removal at 0.4 s. Right, non-additivity between the full-input response and the summed responses to the history and current input. **j**, History-conditioned third-bit effect $\Delta_3(h)$, that is, the change in the mean evidence for odd parity when the third bit is switched from 0 to 1, for each history $h = b_1 b_2$. Each difference compares two patterns containing 100 samples each. Error bars denote one standard error of the difference between their means. **k**, Mean normalized evidence for each input pattern in the intact and $k_{\mathrm{cubic}} = 0$ models (100 samples per pattern). Error bars denote the standard error of the mean. **l**, Accuracy under intact dynamics, $k_{\mathrm{cubic}} = 0$, and segment-wise state reset. All interventions are applied during inference without retraining.

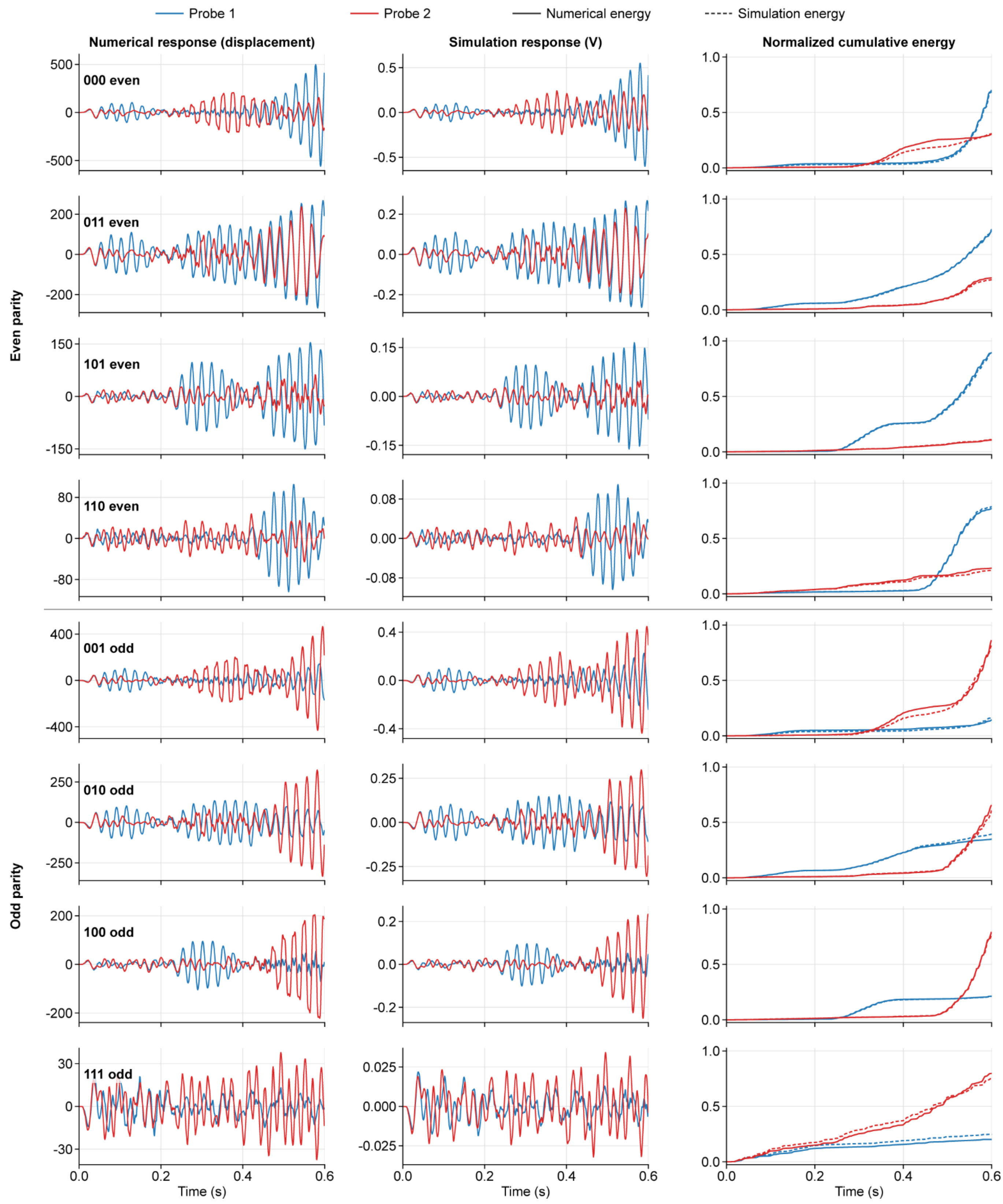

**Extended Data Figure 4. Circuit simulation reproduces the dynamics of the trained nonlinear R$^2$NN.** One test waveform of each three-bit pattern, grouped by parity (even, top four rows; odd, bottom four rows), is applied to the trained nonlinear mechanical R$^2$NN and to its LTspice metacircuit implementation. Left, probe responses of the mechanical model (displacement, normalized units); middle, simulated probe voltages of the metacircuit; right, normalized cumulative probe energies of the mechanical model (solid) and the circuit simulation (dashed). Blue, probe 1; red, probe 2.